\documentclass[11pt,a4paper]{article}

\usepackage[T1]{fontenc}
\usepackage[utf8]{inputenc}
\usepackage{graphicx}
\usepackage{amsmath}
\usepackage{amssymb}
\usepackage{listings}
\usepackage{xcolor}
\usepackage{booktabs}
\usepackage{tabularx}
\usepackage{array}
\usepackage[margin=1in]{geometry}
\usepackage{float}
\usepackage{hyperref}
\usepackage{tikz}
\usetikzlibrary{positioning,arrows.meta,calc,fit,backgrounds}

\definecolor{codegray}{rgb}{0.5,0.5,0.5}
\definecolor{codepurple}{rgb}{0.58,0,0.82}
\definecolor{backcolour}{rgb}{0.95,0.95,0.92}

\lstdefinestyle{mystyle}{
	backgroundcolor=\color{backcolour},
	numberstyle=\color{codegray},
	stringstyle=\color{codepurple},
	basicstyle=\ttfamily\small,
	breakatwhitespace=false,
	breaklines=true,
	captionpos=b,
	keepspaces=true,
	numbers=left,
	numbersep=5pt,
	showspaces=false,
	showstringspaces=false,
	showtabs=false,
	tabsize=2,
	morestring=[b]"
}

\newif\ifshowrev
\showrevfalse
\ifshowrev\definecolor{revE}{rgb}{0.85,0.10,0.45}\else\colorlet{revE}{black}\fi
\newcommand{\revtag}[1]{\ifshowrev\textsuperscript{\textcolor{revE}{\tiny\textsf{[#1]}}}\fi}
\newcommand{\rev}[2]{\textcolor{revE}{#2}\revtag{#1}}

\newenvironment{revblock}[1]{\color{revE}}{}

\newcommand{\code}[1]{\texttt{#1}}
\newcolumntype{Y}{>{\raggedright\arraybackslash}X}
\title{Agent Status Protocol (ASP): A Lightweight Specification for State Reporting in Multi-Agent Systems}

\author{
	\href{https://orcid.org/0009-0008-7513-1255}{\includegraphics[scale=0.06]{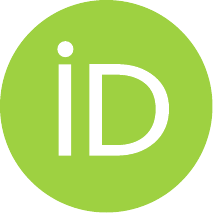}\hspace{1mm}Diego Gosmar}\\
	Open Voice Interoperability Initiative\\
	Linux Foundation AI \& Data\\
	Turin, Italy\\
	\texttt{diego.gosmar@ieee.org}
	\and
	\href{https://orcid.org/0000-0002-3389-2784}{\includegraphics[scale=0.06]{orcid.pdf}\hspace{1mm}Deborah A. Dahl}\\
	Conversational Technologies\\
	Open Voice Interoperability Initiative\\
	Linux Foundation AI \& Data\\
	Plymouth Meeting, PA, USA\\
	\texttt{dahl@conversational-technologies.com}
	\and
	Emmett Coin\\
	ejTalk, Inc.\\
	Open Voice Interoperability Initiative\\
	Linux Foundation AI \& Data\\
	Belleville, Michigan, USA\\
	\texttt{emmett@ejtalk.com}
	\and
	Csaba Bolyós\\
	Open Voice Interoperability Initiative\\
	Linux Foundation AI \& Data\\
	Bratislava, Slovakia\\
	\texttt{bladeszasza@gmail.com}
	\and
	Andreas Zettl\\
	Open Voice Interoperability Initiative\\
	Linux Foundation AI \& Data\\
	Berlin, Germany\\
	\texttt{info@azettl.net}
}

\begin{document}

\maketitle

\begin{abstract}
The Agent Status Protocol (ASP) defines a lightweight way for heterogeneous AI agents to report their operational state to a status server, and for observers such as dashboards to receive that state as soon as it changes. Agents push small JSON events over HTTP POST; the server exposes the aggregated state to observers over Server-Sent Events (SSE). Because an agent that has crashed, hung or been compromised cannot be trusted to report its own failure, ASP does not rely on self-reports alone: the server infers liveness from missed heartbeats, can detect agents that are alive but no longer making progress, and accepts observations from supervisors, guardrails, operators and peers, some of which take precedence over what the agent says about itself while others are shown alongside it. ASP models state along independent dimensions (lifecycle, health, task, floor, governance), so that an agent can be, for example, online, degraded, working on a task and waiting for the floor at the same time. It reuses existing vocabularies: task states from the Agent2Agent (A2A) protocol, floor states from the Open Floor Protocol (OFP), and attribute conventions from the OpenTelemetry GenAI semantic conventions. \rev{P}{This paper presents version 0.1 of ASP as a proposal, not as an approved standard. It is a working draft intended to open a discussion, and the authors invite comments, alternative proposals and independent implementations from the community.}

\end{abstract}


\begin{revblock}{P}
\medskip
\noindent\textbf{Status of this document.}\revtag{P} This document is a proposal submitted for discussion. It has not been reviewed or endorsed by any standards body, and nothing in it should be read as an adopted standard. Its content, including the state names, the message formats and the normative requirements, may change substantially in response to feedback. Comments, objections and alternative proposals are welcome\rev{GH}{. They can be submitted as issues or pull requests in the project's public repository, \url{https://github.com/AgentStatusProtocol/spec}~\cite{aspgithub}, which hosts a summary of the proposal, a draft JSON Schema, example events and illustrative implementations, or} sent to the authors.
\end{revblock}

\section{Introduction}

\subsection{Motivation}
\label{sec:motivation}

AI agents are moving from single assistants to systems in which many agents plan, delegate, debate and call tools on each other's behalf. Each agent is an autonomous and non-deterministic process that depends on external model providers, tools and budgets, any of which can fail, slow down or run out while the agent is still nominally running. One agent can be watched by hand, by reading its logs or its console. A crowd of agents cannot.

The difficulty grows faster than the number of agents. In a fully connected topology of $n$ agents there are $n(n-1)/2$ possible pairwise links: 15 among 6 agents, 190 among 20 and 1{,}225 among 50. Each link is a place where one agent may wait on another whose state it cannot see, and each added agent adds potential links to every existing one. Empirical studies of multi-agent LLM systems confirm that a large share of failures arise from inter-agent misalignment and from tasks that are not properly verified or terminated, and not only from the limits of individual models~\cite{cemri2025mast}.

Without a shared view of agent state, these failures remain blind spots. A crashed or hung agent looks just like a slow one. Peers keep calling a stuck agent, unaware of its state. Rate limits and exhausted quotas surface as poor answers rather than as operational faults. A looping agent can hold the floor of a conversation or consume budget for minutes before anyone notices. An agent that has been compromised, for example by a prompt injection, keeps talking to the others as if nothing had happened.

Three related capabilities are needed to close these gaps. \emph{Tracing} records what happened and when, so that incidents can be reconstructed. \emph{Monitoring} shows live which agents are alive, busy, waiting or blocked, so that health and task problems are seen before users feel them. \emph{Governing} lets an operator or a guardrail act on that picture, for example by quarantining one agent so that the others can route around it, and leaves a timestamped history for audit and post-mortem analysis.

Multi-agent systems increasingly mix agents built on different frameworks and protocols. Each exposes its state differently, or not at all. Operators have no common way to answer basic questions: which agents are alive, what each is doing, who is waiting on whom, and which agent currently holds the floor in each conversation.

ASP provides that common layer. It is deliberately lightweight to adopt, so that any agent able to send an HTTP request can conform.

\subsection{Application Scenarios}
\label{sec:applications}

Live agent state is useful wherever several agents share work and a human or a component is accountable for the outcome. The following scenarios illustrate the range.

\emph{Enterprise agent fleets.} Organisations deploy dozens of specialised agents for tasks such as order handling, document processing or planning, often built by different teams on different frameworks. A shared state view tells operators which agents are degraded, rate-limited or waiting for approval, and lets orchestrators route new work away from them.

\emph{Multi-party conversations and voice.} In open-floor conversations, such as multi-agent debates, contact-centre interactions or voice conferences in which several agents take turns~\cite{gosmar2024aimultiagentinteroperabilityextension}, who holds the floor and who is queued for it is operational information. A convener that stops yielding, or a participant that never gets the floor, becomes visible immediately.

\emph{Safety and guardrail pipelines.} Pipelines in which specialised agents review, correct or filter the output of others, as in hallucination mitigation~\cite{icaart26} and prompt injection mitigation~\cite{gosmar2026promptinjectionmitigationagentic}, need a common way for a guardrail to flag an agent and for every other component to see that flag. The same holds for architectures in which a network of Sentinel Agents monitors inter-agent communication and a Coordinator Agent isolates or quarantines misbehaving agents on the basis of their alerts~\cite{gosmar2025sentinelagentssecuretrustworthy}; Section~\ref{sec:sentinel} shows how ASP supports this pattern.

\emph{Cost and capacity control.} Token budgets, provider quotas and concurrency limits are shared resources. Reporting \code{quota-exhausted}, \code{rate-limited} or queue depth as state lets operators and autoscalers react before a budget overrun or a cascade of retries.

\emph{Regulated and high-assurance deployments.} Regulations such as the EU Artificial Intelligence Act require record-keeping and effective human oversight for high-risk AI systems~\cite{euaiact}. An append-only history of state transitions, together with explicit governance states such as \code{under-review}, gives operators evidence of what each agent was doing and of who intervened, without recording the content of the agents' exchanges.

\subsection{Why an Independent Standard}
\label{sec:whystandard}

Operating deployed enterprise agent fleets at scale requires a reliable and uniform way of monitoring the state of every agent, whoever built it and whatever framework it runs on. Most agent frameworks and observability tools already expose some notion of agent status, but each defines its own. Without a standard, every pair of framework and monitoring tool needs its own integration, the same word means different things in different teams, and a dashboard can show only the agents built with the framework it was designed for. With $F$ frameworks and $T$ tools, bespoke integrations grow as $F \times T$; a shared protocol reduces them to $F + T$. ASP is designed around six properties that a proprietary or framework-bound solution does not provide together.

\emph{Any framework.} Any agent that can send an HTTP POST can report its state. ASP requires no SDK, runtime or vendor, so it does not lock a deployment into one ecosystem.

\emph{Any topology.} The same states describe a floor-managed debate under OFP, a task delegation under A2A and a voice conference, and they apply equally to agents that use neither protocol.

\emph{State only.} Utterances, prompts and replies never enter ASP (Section~\ref{sec:security}). Monitoring therefore does not widen the exposure of what agents say, and the status server does not become a second copy of sensitive content.

\emph{One vocabulary.} \code{idle}, \code{degraded} or \code{awaiting-approval} mean the same thing for every team and tool, so one dashboard can read every agent.

\emph{Server-side liveness.} A missed heartbeat marks an agent as \code{unreachable}. The server decides, so an agent that has died cannot report itself as healthy (Section~\ref{sec:liveness}).

\emph{Governance in the protocol.} Quarantine and policy states are part of the state model, and a state set by a guardrail or an operator takes precedence over the agent's self-report (Section~\ref{sec:security}).

ASP defines the state. It does not define the agent, the framework or the vendor. \rev{P}{This paper proposes ASP as a candidate for such a standard; whether it, or a successor shaped by community input, becomes one depends on review, implementation and consensus beyond its authors.}

\subsection{Why Not an Existing Health Protocol}
\label{sec:whynotexisting}

Web servers and other services already report their state through established mechanisms, such as the liveness and readiness probes of container orchestrators~\cite{k8sprobes} and the gRPC health checking protocol~\cite{grpchealth}. These mechanisms answer one question, whether a process is alive and ready to serve requests, typically with a small set of values such as serving or not serving, and they are usually polled by the infrastructure. Agents need more than that. An agent can be alive and ready while its task waits legitimately for human approval or for input from another agent; it can hold or wait for the floor of a conversation; it can fail because an upstream model provider has rate-limited it or its token budget is exhausted, while the process itself is healthy; and it can be quarantined by a guardrail while still running. None of these conditions can be expressed in a binary health check, and none is observable by probing the process. ASP does not replace existing health protocols, which remain appropriate for the processes that host agents. It reuses what fits from them and from related standards, namely heartbeats with timeouts, HTTP, Server-Sent Events and the OpenTelemetry conventions, and adds the agent-specific dimensions they lack (Section~\ref{sec:related}).

\subsection{Why a State Layer on Top of OpenTelemetry}
\label{sec:whynototel}

OpenTelemetry is the obvious candidate for carrying agent state, and ASP is designed to work with it rather than around it. OpenTelemetry defines how telemetry is produced, carried and exported as metrics, logs and traces, and its GenAI semantic conventions~\cite{otelgenai} name model calls, tool executions and agent invocations, on which the major instrumentation libraries are converging, where necessary through normalisation in the Collector. A deployment could therefore represent agent state with existing OpenTelemetry signals, for example a gauge or a log record carrying a state attribute. What OpenTelemetry does not provide, at the time of writing, are four properties that ASP specifies normatively. None of them concerns transport; all of them concern meaning and behaviour.

\paragraph{A shared vocabulary of agent states.} The GenAI conventions describe what an agent did, not the condition it is in. Without agreed values, \code{degraded} or \code{waiting} in one framework means something different in another, which reproduces for agent state the fragmentation that semantic conventions resolved for model calls. ASP supplies the missing vocabulary, which can be carried as attribute values in any OpenTelemetry signal (Section~\ref{sec:layers}).

\paragraph{A meaning for the absence of a signal.} Telemetry pipelines are designed to tolerate loss through sampling, batching and periodic export. A missing span may mean that nothing happened, that the span was sampled out, or that the agent has died, and whether and when missing telemetry indicates failure is left to the alerting rules of each backend. ASP makes this part of the contract: each agent declares its heartbeat interval, and every conforming server derives \code{unreachable} and \code{stalled} from the same rules (Section~\ref{sec:liveness}).

\paragraph{Every transition, in order.} A periodically exported metric reports one value per export interval, so a state held for a few seconds between two exports is never observed, and spans are exported only when an operation completes. ASP requires every transition to be reported as an event with a per-sender sequence number and kept in an append-only history, which both live control and audit require.

\paragraph{Authority over state.} OpenTelemetry attributes each record to the resource that emitted it. It has no notion that a guardrail may assert the state of another agent, or that such an assertion takes precedence over the agent's self-report. Investigations of agent incidents show why the distinction matters: a record of who claimed to act is not a record of who was observed acting. ASP records the source of every assertion and defines which source prevails (Section~\ref{sec:observation}).

ASP is therefore best understood as a layer of state semantics, liveness rules and authority on top of existing telemetry, not as an alternative to it. Section~\ref{sec:otel} gives the current mapping from OpenTelemetry to ASP. A binding in the opposite direction, in which ASP events are emitted as OpenTelemetry log records linked to the active trace, and an implementation of the status server functions as a component of the OpenTelemetry Collector, are natural extensions of this draft.

\subsection{Why Push State Instead of Asking the Agent}
\label{sec:whypush}

A simpler design would let each observer ask each agent for its state. ASP does not rule this out, and the pull mode of Section~\ref{sec:arch} uses it for agents that cannot be modified, but asking alone does not work well for large, heterogeneous deployments. First, an agent that has crashed or lost connectivity cannot answer, so failure detection has to rely on the absence of an expected signal rather than on a reply; heartbeats provide that signal. Second, asking requires every observer to know and reach every agent: with $N$ agents and $M$ observers polling, traffic grows as $N \times M$, and agents behind NAT or firewalls would need inbound ports. Third, some states are not the agent's to report: \code{unreachable} is inferred by the server, and governance states may be set by a guardrail or an operator against the agent's own report. Fourth, a server that receives every transition can keep a complete history for audit, whereas periodic polling misses transitions that happen between two polls. Finally, answers are only comparable if agents use the same names, so a shared state vocabulary is needed however the state is obtained, which is why it can be adopted on its own (Section~\ref{sec:layers}).

\subsection{Scope}

ASP covers agent registration and deregistration with a status server, periodic liveness signalling (heartbeat), reporting of state transitions along defined dimensions, and live distribution of aggregated state to observers.

\subsection{Non-goals}

ASP does not define how agents communicate with each other, which is the role of A2A~\cite{a2a}, OFP~\cite{ofp} and similar protocols. It does not define task payloads, prompts or model outputs, nor full distributed tracing, which is the role of OpenTelemetry~\cite{otelgenai}. Agent discovery and capability negotiation are also out of scope.

\subsection{Relationship to Existing Standards}
\label{sec:relstd}

ASP is designed as a profile, not a replacement. Task states align with A2A. Floor and role states align with OFP. Metric and identity attributes align with the OpenTelemetry GenAI semantic conventions, so that an ASP server can also ingest OTLP data. Section~\ref{sec:mappings} gives the mappings. ASP is independent of each of these protocols: it observes agents that speak them but does not extend or depend on any of them, and an agent that speaks none of them can still conform.

\subsection{Specification Layers}
\label{sec:layers}

This specification is organised in three layers that can be adopted separately. The \emph{state vocabulary} (Section~\ref{sec:states}) defines the dimensions and the standard state names; it does not depend on any transport and MAY be used on its own, for example as attribute values in logs, in OpenTelemetry spans or in the metadata of another agent protocol. The \emph{event layer} (Section~\ref{sec:messages}) defines the JSON messages that carry state transitions. The \emph{behaviour layer} (Sections~\ref{sec:liveness}, \ref{sec:transport} and~\ref{sec:security}) defines what a status server does with those messages: liveness inference, transport binding and security rules. An implementation MAY conform to the state vocabulary alone, to the state vocabulary and the event layer, or to all three layers. For this reason the state model is presented before the message types.

\section{Related Work}
\label{sec:related}

ASP draws on four bodies of work: agent communication protocols, lifecycle management in classical multi-agent platforms, observability of LLM agents, and failure detection in distributed systems. It also builds on earlier work on multi-party agent conversations and their safety applications. This section reviews each and then summarises what ASP adds (Table~\ref{tab:related}).

\subsection{Agent Communication Protocols}

Agent communication has a long history, from KQML and FIPA ACL to the protocols that have emerged for LLM-based agents. The Model Context Protocol (MCP)~\cite{mcp} connects an agent to tools and data sources; the Agent2Agent protocol (A2A)~\cite{a2a} lets agents delegate tasks to each other and describes their capabilities in Agent Cards; the Agent Communication Protocol (ACP) and the Agent Network Protocol (ANP) address REST-based messaging and open-network discovery respectively; a recent survey compares these four protocols~\cite{ehtesham2025survey}. The Open Floor Protocol (OFP)~\cite{ofp} manages multi-party conversations in which a convener grants and revokes the floor.

These protocols define how agents talk, and some expose part of an agent's state as a side effect. A2A defines task states, but they are visible only to the client of each task and say nothing about whether the agent itself is healthy or alive. An Agent Card is a static description rather than a live status. OFP floor events describe one conversation to its participants. MCP describes tools, not agents. None of them defines liveness, health or governance, and none gives a third-party observer an aggregated, live view of all agents. ASP complements these protocols rather than competing with them, and reuses their vocabularies where they exist (Section~\ref{sec:mappings}).

\subsection{Lifecycle Management in Classical Multi-Agent Platforms}

The FIPA Agent Management Specification~\cite{fipaam} defined an agent platform in which an Agent Management System keeps a directory of agents and a lifecycle with states such as initiated, active, suspended, waiting and transit. Platforms such as JADE~\cite{bellifemine2007jade} implemented this model. ASP inherits the idea of an explicit lifecycle. It differs in three respects: the FIPA lifecycle lives inside a single agent platform rather than across heterogeneous frameworks; it has a single dimension, so an agent cannot be simultaneously active, degraded and waiting for the floor; and it predates the failure modes typical of LLM agents, such as rate limits, exhausted quotas, unavailable tool servers and suspected prompt injection.

\subsection{Observability of LLM Agents}

The OpenTelemetry GenAI semantic conventions~\cite{otelgenai} standardise span and metric attributes for model calls and agent invocations, and the AgentOps taxonomy~\cite{dong2024agentops} catalogues the artefacts that should be traced across the lifecycle of LLM agents. Commercial and open-source tracing platforms build on these ideas. This work is essential for debugging and evaluation, but it is organised around traces of past operations: spans are exported once an operation completes, they often carry prompts and model outputs, and there is no model of the current state of an agent or of its liveness between operations. ASP is complementary: it adopts the OpenTelemetry identifiers and can ingest OTLP data (Section~\ref{sec:otel}), and it adds the live, content-free state that traces do not express.

\subsection{Failure Detection and Health Checking}

Distributed systems have long detected crashed processes with heartbeats and timeouts, formalised as unreliable failure detectors~\cite{chandra1996failure} and refined by accrual detectors that adapt to network conditions~\cite{hayashibara2004phi}. Container orchestrators check the liveness and readiness of processes through periodic probes~\cite{k8sprobes}, and gRPC defines a standard health checking service~\cite{grpchealth}. ASP applies the same heartbeat-and-timeout principle (Section~\ref{sec:liveness}), with two differences: liveness is inferred from heartbeats the agent pushes, rather than from probes the orchestrator sends, and a process that is alive is only the first dimension of an agent's state, alongside health, tasks, floor and governance.

\subsection{Failures, Guardrails and Multi-Party Conversations}

Empirical work on multi-agent LLM systems has produced taxonomies of how they fail, including inter-agent misalignment and failures of termination and verification~\cite{cemri2025mast}. Guardrail toolkits such as NeMo Guardrails~\cite{rebedea2023nemo} constrain what individual agents may say or do. ASP builds directly on earlier work on multi-party agent conversations: the multi-agent interoperability extension for managing multiparty conversations~\cite{gosmar2024aimultiagentinteroperabilityextension} introduced the convener and floor concepts that ASP adopts in its floor dimension, and the same open-floor approach has been applied to hallucination mitigation in multi-agent pipelines~\cite{icaart26} and to prompt injection mitigation, where specialised agents detect and neutralise injected instructions~\cite{gosmar2026promptinjectionmitigationagentic}. Sentinel Agents~\cite{gosmar2025sentinelagentssecuretrustworthy} generalise this idea into a distributed security layer: sentinels combine LLM-based semantic analysis, behavioural analytics, retrieval-augmented verification and cross-agent anomaly detection to oversee inter-agent communication, while a Coordinator Agent ingests their alerts, adapts policies and isolates or quarantines misbehaving agents. That work defines who detects threats and who decides; it leaves open the format in which alerts and decisions are exchanged and made visible to the rest of the system, which is the gap ASP fills. More generally, this line of work shows that the behaviour of a multi-agent pipeline depends on the state of each agent and on who holds the floor at any moment, but none of it defines a common runtime signal through which that state becomes visible to the rest of the system.

\subsection{What ASP Adds}

Table~\ref{tab:related} compares ASP with the closest existing approaches. The comparison concerns the exposure of agent state to observers, not the broader purpose of each approach, most of which ASP is designed to complement. ASP contributes:

\begin{itemize}
\item a multi-dimensional state model, in which lifecycle, health, task, floor and governance vary independently, so that an agent can be online, degraded, working and waiting for the floor at the same time;
\item liveness inferred by the server from heartbeats declared by each agent, so that an agent that has died is marked \code{unreachable} rather than trusted to report itself;
\item governance states in the protocol, with precedence for guardrails and operators over self-reports and a record of who set each state;
\item a profile that reuses A2A task states, OFP floor states and OpenTelemetry identifiers, with explicit mappings that let adapters cover agents that cannot be modified;
\item a content-free design, in which events carry state but never prompts, outputs or user data;
\item a layered specification whose state vocabulary can be adopted without the event and behaviour layers (Section~\ref{sec:layers}).
\end{itemize}

\begin{table}[H]
\centering
\caption{Exposure of agent state to observers in existing approaches and in ASP. \checkmark: supported; $\circ$: partial or limited scope; --: not addressed.}
\label{tab:related}
\small
\begin{tabularx}{\textwidth}{@{}Ycccccc@{}}
\toprule
Approach & Lifecycle/health & Liveness & Task & Floor & Governance & Aggregated live view \\
\midrule
FIPA Agent Management~\cite{fipaam} & $\circ$ & -- & -- & -- & -- & $\circ$ \\
A2A~\cite{a2a} & -- & -- & \checkmark & -- & -- & $\circ$ \\
OFP~\cite{ofp} & -- & -- & -- & \checkmark & -- & $\circ$ \\
MCP~\cite{mcp} & -- & -- & -- & -- & -- & -- \\
OpenTelemetry GenAI~\cite{otelgenai} & -- & -- & $\circ$ & -- & -- & -- \\
Health probes~\cite{k8sprobes,grpchealth} & $\circ$ & \checkmark & -- & -- & -- & $\circ$ \\
ASP & \checkmark & \checkmark & \checkmark & \checkmark & \checkmark & \checkmark \\
\bottomrule
\end{tabularx}
\end{table}

\section{Terminology}

The key words MUST, MUST NOT, SHOULD, SHOULD NOT and MAY are to be interpreted as described in RFC~2119~\cite{rfc2119} and RFC~8174~\cite{rfc8174}. \rev{P}{In this proposal they state the requirements that a conforming implementation would have to meet if ASP were adopted; they do not imply that any standards body has endorsed these requirements.} Table~\ref{tab:terms} defines the terms used in this specification.

\begin{table}[H]
\centering
\caption{Terminology.}
\label{tab:terms}
\small
\begin{tabularx}{\textwidth}{@{}lY@{}}
\toprule
Term & Definition \\
\midrule
Agent & An autonomous software component that uses AI models, memory and tools to pursue goals, make decisions or act on behalf of users or other agents, and whose state is reported using ASP, either natively or through an adapter. \\
Status server & The component that receives ASP events, maintains current state and distributes it. \rev{SV}{In the reference binding it is an HTTP service run by the operator of a deployment of agents, but its functions may also be embedded in an existing component (Section~\ref{sec:topologies});} its scope is \rev{SV}{one} deployment, not the Internet (Section~\ref{sec:placement}). \\
Observer & Any client that consumes aggregated state. The typical observer is a dashboard used by a human operator; other observers are software, such as alerting, autoscaling or orchestration components. \\
Operator & The human, or the software acting on a human's behalf (for example an orchestrator, supervisor or autoscaler), responsible for running a deployment of agents. Operator actions such as pausing an agent or setting a governance state reach the status server through a credential authorised for that purpose (Section~\ref{sec:security}). \\
Event & A single JSON message about one agent, sent to the status server by that agent or by a component authorised to report on it, or generated by the server itself. \\
Dimension & An independent axis of state (lifecycle, health, task, floor, governance). \\
State & A value within one dimension at a given time. \\
Adapter & A component that emits ASP events on behalf of an agent that does not implement ASP natively. \\
\bottomrule
\end{tabularx}
\end{table}

\section{Architecture}
\label{sec:arch}

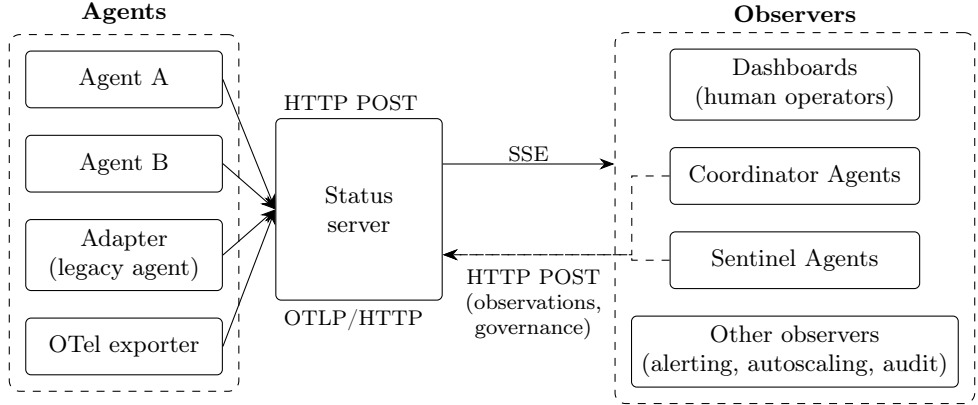
\begin{figure}[t]
\centering
\begin{tikzpicture}[
  node distance=0.35cm and 1.6cm,
  box/.style={draw, rounded corners=2pt, minimum width=2.6cm, minimum height=0.75cm, align=center, font=\footnotesize},
  obox/.style={box, minimum width=3.3cm},
  grp/.style={draw, dashed, rounded corners=4pt, inner sep=6pt},
  gl/.style={font=\footnotesize\bfseries, anchor=south},
  lbl/.style={font=\scriptsize, fill=white, inner sep=1pt},
  >={Stealth[length=2mm]}
]
\node[box] (a1) {Agent A};
\node[box, below=of a1] (a2) {Agent B};
\node[box, below=of a2] (ad) {Adapter\\(legacy agent)};
\node[box, below=of ad] (ot) {OTel exporter};
\node[box, minimum width=2.2cm, minimum height=2.4cm, right=2.0cm of $(a2)!0.5!(ad)$] (s) {Status\\server};
\node[obox, right=3.0cm of s, yshift=1.65cm] (o1) {Dashboards\\(human operators)};
\node[obox, below=of o1] (o2) {Coordinator Agents};
\node[obox, below=of o2] (o3) {Sentinel Agents};
\node[obox, below=of o3] (o4) {Other observers\\(alerting, autoscaling, audit)};
\begin{scope}[on background layer]
\node[grp, fit=(a1)(ot)] (ga) {};
\node[grp, fit=(o1)(o4)] (go) {};
\end{scope}
\node[gl] at (ga.north) {Agents};
\node[gl] at (go.north) {Observers};
\draw[->] (a1.east) -- (s.west);
\draw[->] (a2.east) -- (s.west);
\draw[->] (ad.east) -- (s.west);
\draw[->] (ot.east) -- (s.west);
\node[lbl, anchor=south west] at ([xshift=2pt,yshift=2pt]s.north west) {HTTP POST};
\node[lbl, anchor=north west] at ([xshift=2pt,yshift=-2pt]s.south west) {OTLP/HTTP};
\draw[->] ([yshift=0.6cm]s.east) -- ([yshift=0.6cm]s.east -| go.west) node[lbl, midway, above] {SSE};
\coordinate (fb) at ([yshift=-0.6cm]s.east);
\draw[->, dashed] (o2.west) -- ++(-0.5cm,0) |- (fb);
\draw[->, dashed] (o3.west) -- ++(-0.5cm,0) |- (fb);
\node[lbl, align=center, anchor=north] at ([xshift=1.2cm,yshift=-0.15cm]fb) {HTTP POST\\(observations,\\governance)};
\end{tikzpicture}
\caption{ASP architecture. Agents and adapters push events; observers receive updates over SSE. Observers include dashboards for human operators and software components; some of them, such as Coordinator Agents or Sentinel Agents, also report on other agents (dashed arrow).}
\label{fig:arch}
\end{figure}

ASP has three roles: agents push events, the status server aggregates them, and observers subscribe to the result (Fig.~\ref{fig:arch}).

Agents initiate every connection, so agents behind NAT or firewalls work without inbound ports; the only exception is the OPTIONAL active probe of Section~\ref{sec:probe}. Observers receive updates as soon as the server processes an event; no polling is required.

The status server does not connect to an agent, except for the OPTIONAL active probe. Each agent, or its adapter, is configured at deployment time with the base URL of the status server and a credential, for example through environment variables or a configuration file, and sends \code{register} before its first event to that server (Section~\ref{sec:register}). How that configuration is distributed is a deployment concern, and discovery of status servers is out of scope. Agents that cannot be configured in this way are covered by adapters, or by the pull mode described below.

A status server MAY also operate in pull mode for agents it cannot modify, for example by periodically reading an A2A Agent Card or a health endpoint and synthesising ASP events. Pull mode is outside the core specification. Appendix~A shows an illustrative observer dashboard.

\subsection{Observers}
\label{sec:observers}

An observer is any client that reads the aggregated state from the status server through the distribution endpoints: it fetches a snapshot of current state, subscribes to the SSE stream, receives each change as soon as the server processes it, and can reconnect without losing updates (Section~\ref{sec:transport}). An observer does not need to contact any agent to learn its state. ASP does not prevent it from doing so: a component may also query agents through other protocols, for example by requesting the status of an A2A task or by asking an agent to justify a claim during verification, and MAY report the outcome to the status server as an observation with a \code{source} other than \code{self} (Section~\ref{sec:observation}). Observers need authorisation, may filter the stream by agent or conversation, and see only state, never the content of what agents say.

Observers fall into three broad groups, shown in Fig.~\ref{fig:arch}. The typical observer is a dashboard through which human operators supervise a deployment, such as the one in Appendix~A. Software observers act on the same stream without a human in the loop: alerting systems notify on-call staff when an agent becomes \code{unreachable}, autoscalers read queue depth and active tasks, orchestrators stop routing work to agents that are \code{degraded} or \code{quarantined}, and exporters feed audit archives or security monitoring. Finally, some observers are themselves agents. A Coordinator Agent reads the stream to receive alerts and to correlate them with the health and task states of the agents concerned before deciding on governance actions. Sentinel Agents may read it to use state patterns as an additional signal and to focus their scrutiny (Section~\ref{sec:sentinel}); a sentinel that only reports alerts, without consuming state, is not an observer.

Being an observer is a function, not a kind of component. A single component may register and report its own state as an agent, read the stream as an observer, and report on other agents with a \code{source} other than \code{self}; a Coordinator Agent typically does all three. The same component may also interact with agents through channels outside ASP: a Sentinel Agent inspects inter-agent messages through the conversation protocol and may query a suspect agent directly, and a Coordinator Agent may query agents and enforce its decisions through the convener or the orchestrator, for example by revoking an agent's floor. ASP carries the resulting state, not those interactions. The term is unrelated to the OFP conversation role \code{observer}, which denotes an agent that listens to a conversation without speaking rights.

\subsection{Status Server Placement}
\label{sec:placement}

\rev{SV}{In the reference binding,} a status server is an ordinary HTTP service. It is not an Internet-wide registry. Each deployment runs its own server, and an agent reports only to the server it is configured for. Several servers MAY coexist, for example one per environment or per business unit; aggregation across servers is out of scope for this draft.

The status server is independent of any conversation protocol. In an OFP deployment it is not part of the floor: the convener is an agent like any other and reports its own state to the server, and the floor manager, where present, is a separate component that MAY do the same. A floor manager MAY run an ASP adapter that reports floor transitions on behalf of the agents in its conversations, and MAY be co-located with the status server, but neither is required.

\begin{revblock}{SV}
\subsection{Deployment Topologies}
\label{sec:topologies}

\revtag{SV}ASP does not require a dedicated server, but it does require that the functions of a status server be performed by some component: inferring liveness, including the states an agent cannot report about itself; enforcing the precedence of governance states over self-reports; checking credentials and ordering events; and distributing state to observers while keeping a history for audit. The HTTP and SSE binding of Section~\ref{sec:transport} is the reference realisation of these functions, but other topologies are possible. The functions may be \emph{embedded} in a component that already exists, such as an orchestrator, an agent runtime or an OFP convener, without deploying a separate service. They may rely on a \emph{message broker}, for example MQTT, whose retained messages and Last Will and Testament can provide a snapshot of state and part of liveness detection. Or events may travel \emph{in-band} inside the conversation or task protocol, with each participant building its own view. The more these functions are distributed, the weaker liveness and governance become, because no single component has an authoritative view. A deployment MAY choose any topology, but it SHOULD identify which component performs each function; where no component checks credentials at ingest, events SHOULD be signed (Section~\ref{sec:security}).
\end{revblock}

\begin{revblock}{SL}
\subsection{Stateful and Stateless Agents}
\label{sec:stateless}

\revtag{SL}ASP works best with agents that keep their own operational state: a stable identity, a process that keeps running between requests, and knowledge of the tasks and conversations they are engaged in. Stateless agents that exist only for the duration of a single invocation, for example as serverless functions, can also be represented, but not natively: between invocations no process sends heartbeats, several short-lived instances may serve the same logical agent, and states that span several invocations are not retained. For such agents it is RECOMMENDED that a component that persists across invocations, such as the orchestrator or the hosting platform, report on their behalf as an adapter, using the \code{agentId} of the logical agent. The distinction concerns operational state, not conversational memory: an agent that remembers nothing of previous conversations but runs as a long-lived service can report its state natively.
\end{revblock}

\section{State Model}
\label{sec:states}

This section is the state vocabulary layer (Section~\ref{sec:layers}) and can be used independently of the event and behaviour layers. An agent's state is a tuple of values, one per dimension. Dimensions are independent: a change in one does not imply a change in another, except where Section~\ref{sec:liveness} states otherwise. Table~\ref{tab:dimensions} summarises the dimensions.

\begin{table}[H]
\centering
\caption{State dimensions.}
\label{tab:dimensions}
\small
\begin{tabularx}{\textwidth}{@{}llY@{}}
\toprule
Dimension & Cardinality & Reported via \\
\midrule
lifecycle & one per agent & \code{state\_change}, \code{heartbeat} \\
health & one per agent & \code{state\_change}, \code{heartbeat} \\
task & one per task & \code{state\_change} \\
floor & one per conversation & \code{floor\_change} \\
governance & one per agent & \code{state\_change} \\
\bottomrule
\end{tabularx}
\end{table}

\subsection{Lifecycle}
\label{sec:lifecycle}

\begin{figure}[t]
\centering
\begin{tikzpicture}[
  st/.style={draw, rounded corners=6pt, minimum height=0.6cm, minimum width=1.7cm, font=\scriptsize\ttfamily},
  >={Stealth[length=1.8mm]},
  node distance=0.5cm and 0.55cm
]
\node[st] (reg) at (0,0) {registering};
\node[st] (init) at (2.3,0) {initializing};
\node[st] (on) at (4.6,0) {online};
\node[st] (paused) at (0,-1.3) {paused};
\node[st] (idle) at (2.3,-1.3) {idle};
\node[st] (busy) at (4.6,-1.3) {busy};
\node[st] (drain) at (6.9,-1.3) {draining};
\node[st] (stop) at (6.9,-2.6) {stopping};
\node[st] (off) at (9.2,-2.6) {offline};
\node[st] (cap) at (4.6,-3.6) {at-capacity};
\draw[->] (reg) -- (init);
\draw[->] (init) -- (on);
\draw[->] (on) -- (idle);
\draw[->] (on) -- (busy);
\draw[->] (idle.10) -- (busy.170);
\draw[->] (busy.190) -- (idle.350);
\draw[->] (idle.170) -- (paused.10);
\draw[->] (paused.350) -- (idle.190);
\draw[->] (busy) -- (drain);
\draw[->] (drain) -- (stop);
\draw[->] (idle.south) |- (stop.west);
\draw[->] (stop) -- (off);
\draw[->] (busy.260) -- (cap.100);
\draw[->] (cap.80) -- (busy.280);
\end{tikzpicture}
\caption{Lifecycle transitions. The server-assigned \code{unreachable} and the supervisor-reported \code{crashed} are reachable from any state and are omitted for clarity. An agent in \code{at-capacity} returns to \code{busy} when capacity frees up.}
\label{fig:lifecycle}
\end{figure}
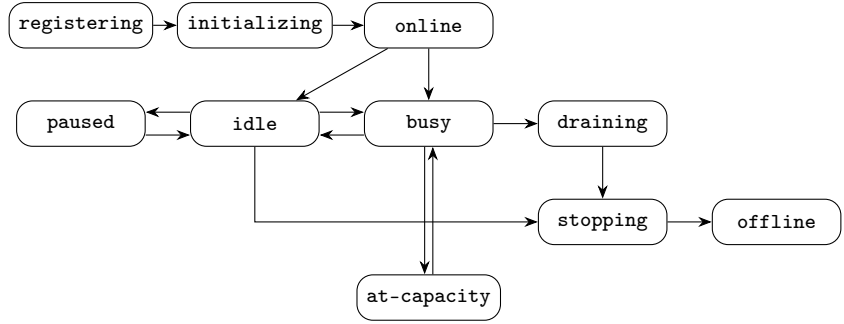

Table~\ref{tab:lifecycle} lists the lifecycle states and Fig.~\ref{fig:lifecycle} the permitted transitions. From any state, the server MAY assign \code{unreachable}, and a supervisor MAY report \code{crashed}. Receiving any event from an \code{unreachable} agent returns it to the state that event reports.

The states \code{idle} and \code{busy} describe whether work is in progress, not whether the agent is at full capacity. An agent working on one task that can still accept more is \code{busy}. Spare capacity is reported through the OPTIONAL \code{heartbeat} metrics \code{activeTasks} and \code{maxConcurrentTasks}. An agent that can accept no further work reports lifecycle \code{at-capacity}: this is a normal operating condition, not a fault, and is distinct from health \code{overloaded}, which signals a request queue above threshold. The states \code{idle}, \code{busy} and \code{at-capacity} are reported by the agent, since only the agent knows its own capacity.

\begin{table}[H]
\centering
\caption{Lifecycle states.}
\label{tab:lifecycle}
\small
\begin{tabularx}{\textwidth}{@{}lY@{}}
\toprule
State & Meaning \\
\midrule
\code{registering} & Announcing itself to the server. \\
\code{initializing} & Loading models, tools and configuration. \\
\code{online} & Active and ready. \\
\code{idle} & Online with no work in progress. \\
\code{busy} & Online with work in progress, and able to accept more. \\
\code{at-capacity} & Online with work in progress and no capacity for further work. \\
\code{paused} & Suspended by an operator; accepts no new work. \\
\code{draining} & Finishing current work, accepting no new work; typically precedes shutdown. \\
\code{stopping} & Shutting down. \\
\code{offline} & Cleanly stopped, after \code{deregister}. \\
\code{unreachable} & Heartbeat missed. Server-assigned only (Section~\ref{sec:liveness}). \\
\code{crashed} & Terminated abnormally, as detected by a supervisor or adapter. \\
\bottomrule
\end{tabularx}
\end{table}

\subsection{Health}

Health transitions are unconstrained: any state MAY move to any other. Table~\ref{tab:health} lists the health states.

\begin{table}[H]
\centering
\caption{Health states.}
\label{tab:health}
\small
\begin{tabularx}{\textwidth}{@{}lY@{}}
\toprule
State & Meaning \\
\midrule
\code{healthy} & Operating normally. \\
\code{degraded} & Working with reduced performance, e.g.\ high latency or fallback to a smaller model. \\
\code{rate-limited} & Temporarily blocked by an upstream provider's rate limit. \\
\code{quota-exhausted} & Token or cost budget used up. \\
\code{dependency-down} & A required tool, API or MCP server is not responding. \\
\code{auth-error} & Credentials expired or invalid. \\
\code{overloaded} & Request queue above threshold. \\
\code{unhealthy} & Unable to perform its function. \\
\code{stalled} & Alive but making no progress on its work. Server-assigned (Section~\ref{sec:progress}). \\
\bottomrule
\end{tabularx}
\end{table}

\subsection{Task}

Task state is tracked per \code{taskId}, which every \code{state\_change} in the task dimension carries (Section~\ref{sec:statechange}); an agent MAY have several tasks open at the same time, each with its own state. Table~\ref{tab:task} lists the task states. States of origin A2A are taken unchanged from the A2A specification~\cite{a2a}; the others are ASP extensions. A task in a terminal state MUST NOT transition further. The server SHOULD remove terminal tasks from the live view after a configurable retention period.

\begin{table}[H]
\centering
\caption{Task states.}
\label{tab:task}
\small
\begin{tabularx}{\textwidth}{@{}llcY@{}}
\toprule
State & Origin & Terminal & Meaning \\
\midrule
\code{submitted} & A2A & no & Received, not yet started. \\
\code{queued} & ASP & no & Waiting behind other tasks. \\
\code{working} & A2A & no & In progress. \\
\code{input-required} & A2A & no & Needs input from a user or another agent. \\
\code{auth-required} & A2A & no & Needs additional authorisation. \\
\code{awaiting-approval} & ASP & no & Waiting for human-in-the-loop approval. \\
\code{delegated} & ASP & no & Part of the work passed to another agent; awaiting its result. \\
\code{retrying} & ASP & no & Repeating a failed step. \\
\code{completed} & A2A & yes & Finished successfully. \\
\code{failed} & A2A & yes & Finished with an error. \\
\code{canceled} & A2A & yes & Cancelled by the requester. \\
\code{rejected} & A2A & yes & Declined by the agent (out of scope, policy). \\
\code{timed-out} & ASP & yes & Exceeded its maximum duration. \\
\code{unknown} & A2A & no & State cannot be determined. \\
\bottomrule
\end{tabularx}
\end{table}

\subsection{Floor}
\label{sec:floor}

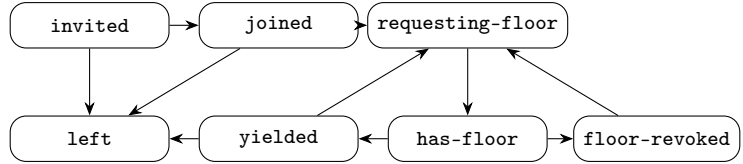
\begin{figure}[t]
\centering
\begin{tikzpicture}[
  st/.style={draw, rounded corners=6pt, minimum height=0.6cm, minimum width=2.1cm, font=\scriptsize\ttfamily},
  >={Stealth[length=1.8mm]},
  node distance=0.6cm and 0.7cm
]
\node[st] (inv) at (0,0) {invited};
\node[st] (join) at (2.5,0) {joined};
\node[st] (req) at (5,0) {requesting-floor};
\node[st] (left) at (0,-1.5) {left};
\node[st] (yld) at (2.5,-1.5) {yielded};
\node[st] (has) at (5,-1.5) {has-floor};
\node[st] (rev) at (7.5,-1.5) {floor-revoked};
\draw[->] (inv) -- (join);
\draw[->] (join) -- (req);
\draw[->] (req) -- (has);
\draw[->] (has) -- (yld);
\draw[->] (has) -- (rev);
\draw[->] (yld) -- (req);
\draw[->] (rev) -- (req);
\draw[->] (join) -- (left);
\draw[->] (yld) -- (left);
\draw[->] (inv) -- (left);
\end{tikzpicture}
\caption{Floor transitions within one conversation. An agent MAY enter the conversation directly in \code{joined} without a prior invitation. An invited agent that declines or is uninvited moves directly to \code{left}.}
\label{fig:floor}
\end{figure}

Floor states are aligned with OFP floor management~\cite{ofp,gosmar2024aimultiagentinteroperabilityextension} and are tracked per \code{conversationId}. Table~\ref{tab:floor} lists them and Fig.~\ref{fig:floor} shows the permitted transitions.

Floor states describe an agent's participation in one conversation, not the agent itself. An agent in \code{left} for a conversation stops sending \code{floor\_change} events for that conversation, but it remains active: it continues to send heartbeats and to report its lifecycle, health, task and governance states, and it MAY take part in other conversations. An agent that stopped reporting altogether after leaving a conversation would be marked \code{unreachable} (Section~\ref{sec:liveness}); an agent that is shutting down sends \code{deregister} instead.

ASP does not define conversation roles. Roles such as convener, participant or observer belong to the conversation protocol, and their meaning is defined there. A \code{floor\_change} event MAY carry the agent's role in the conversation in the OPTIONAL \code{role} field, using the value defined by that protocol, for example \code{convener} or \code{participant} for OFP; the status server stores and forwards the value without interpreting it.

\begin{table}[H]
\centering
\caption{Floor states.}
\label{tab:floor}
\small
\begin{tabularx}{\textwidth}{@{}lY@{}}
\toprule
State & Meaning \\
\midrule
\code{not-joined} & Not part of the conversation. \\
\code{invited} & Invited, not yet joined. \\
\code{joined} & Participating and listening. \\
\code{requesting-floor} & Has asked to speak. \\
\code{has-floor} & Currently speaking. \\
\code{yielded} & Has given up the floor. \\
\code{floor-revoked} & Floor removed by the convener. \\
\code{left} & Has left the conversation, for example after sending \code{bye}, after being removed from it, or after declining the invitation. \\
\bottomrule
\end{tabularx}
\end{table}

\subsection{Governance}

\label{sec:governance}
Governance is an OPTIONAL dimension for production deployments (Table~\ref{tab:governance}). Governance states MAY be set by the agent itself, by a guardrail component, or by an operator through the server. Observers SHOULD display any state other than \code{normal} prominently. The \code{quarantined} state is intended, for example, for agents flagged by a prompt injection detection layer such as the one described in~\cite{gosmar2026promptinjectionmitigationagentic}, or quarantined by a Coordinator Agent on the basis of Sentinel Agent alerts~\cite{gosmar2025sentinelagentssecuretrustworthy} (Section~\ref{sec:sentinel}).

\begin{table}[H]
\centering
\caption{Governance states.}
\label{tab:governance}
\small
\begin{tabularx}{\textwidth}{@{}lY@{}}
\toprule
State & Meaning \\
\midrule
\code{normal} & No governance restriction. \\
\code{sandboxed} & Operating with reduced permissions. \\
\code{quarantined} & Isolated after anomalous behaviour, e.g.\ suspected prompt injection. \\
\code{policy-violation} & Attempted an action forbidden by policy. \\
\code{under-review} & Suspended pending review by an operator, human or software (Table~\ref{tab:terms}). \\
\bottomrule
\end{tabularx}
\end{table}

\section{Message Types}
\label{sec:messages}

ASP defines five event types. Every event is a JSON object sent as the body of an HTTP POST with \code{Content-Type: application/json}.

\subsection{Common Fields}

Table~\ref{tab:common} lists the fields common to all events. The examples in this section omit \code{aspVersion}, \code{eventId} and \code{seq} for brevity.

\begin{table}[H]
\centering
\caption{Common fields of all ASP events.}
\label{tab:common}
\small
\begin{tabularx}{\textwidth}{@{}llcY@{}}
\toprule
Field & Type & Req. & Description \\
\midrule
\code{aspVersion} & string & yes & Protocol version, e.g.\ \code{"0.1"}. \\
\code{type} & string & yes & One of \code{register}, \code{heartbeat}, \code{state\_change}, \code{floor\_change}, \code{deregister}. \\
\code{agentId} & string (URI) & yes & Stable, globally unique agent identifier. A URN is RECOMMENDED. \\
\code{eventId} & string & yes & Unique per event, used for deduplication. UUID RECOMMENDED. \\
\code{seq} & integer & yes & Monotonically increasing per agent and per sender: an agent numbers its own events, and each component that reports on it numbers its own events about that agent. Used for ordering. \\
\code{ts} & string & yes & Event time in RFC~3339~\cite{rfc3339} format, UTC. \\
\code{source} & string & no & Who reports the event: \code{self} (default), \code{supervisor}, \code{guardrail}, \code{operator} or \code{peer}. Events generated by the server carry \code{server}. See Section~\ref{sec:observation}. \\
\bottomrule
\end{tabularx}
\end{table}

\subsection{register}
\label{sec:register}

Sent once, before the agent's first event to a given status server. An agent registers when it begins to operate in the deployment that the server monitors: typically at start-up, or when it is first engaged by that deployment, for example when it is invited to an OFP conversation or receives its first A2A task. Activity of the agent before that moment, unrelated to the deployment, is outside the server's scope. Registration is independent of any conversation: joining a conversation is reported separately as a \code{floor\_change} to \code{joined}. An agent that stops operating in the deployment sends \code{deregister}; an agent that reports to several status servers registers with each. Registration MUST precede any other event from that agent. The field \code{heartbeatIntervalSec} is REQUIRED and MUST be between 1 and 300. The fields \code{healthUrl}, an endpoint the server MAY probe (Section~\ref{sec:probe}), and \code{stallAfterSec}, the time without progress after which the agent is considered stalled (Section~\ref{sec:progress}), are OPTIONAL.

\begin{lstlisting}
{
  "type": "register",
  "agentId": "urn:agent:example:research-assistant",
  "name": "Research Assistant",
  "version": "1.2.0",
  "protocols": ["a2a", "ofp"],
  "capabilities": ["web-research", "summarization"],
  "heartbeatIntervalSec": 10,
  "ts": "2026-09-22T17:00:00Z"
}
\end{lstlisting}

\subsection{heartbeat}
\label{sec:hb}

Sent periodically to prove liveness. It SHOULD be as small as possible. It SHOULD be emitted from the agent's main work loop rather than from an independent thread, and SHOULD carry \code{progressSeq}, a counter that the agent increments whenever it makes progress on its work (Section~\ref{sec:progress}). It MAY carry \code{activity}, a short code that tells observers what a long wait is spent on without revealing its content. Recommended values are \code{model-call}, \code{tool-call}, \code{media-generation}, \code{waiting-floor}, \code{waiting-peer} and \code{waiting-input}; custom values follow Section~\ref{sec:ext}. The field is a hint for observers and does not change the agent's state in any dimension. The \code{metrics} object is OPTIONAL; Section~\ref{sec:otel} lists recommended metric names.

\begin{lstlisting}
{
  "type": "heartbeat",
  "agentId": "urn:agent:example:research-assistant",
  "lifecycle": "busy",
  "health": "healthy",
  "progressSeq": 1742,
  "activity": "model-call",
  "metrics": { "latencyMs": 820, "queueDepth": 2, "activeTasks": 1, "maxConcurrentTasks": 4 },
  "ts": "2026-09-22T17:00:10Z"
}
\end{lstlisting}

\subsection{state\_change}
\label{sec:statechange}

Sent on every transition in any dimension other than floor. The fields \code{dimension}, \code{from} and \code{to} are REQUIRED. The field \code{taskId} is REQUIRED when \code{dimension} is \code{task}. The field \code{maxDurationSec} is OPTIONAL and applies to transitions of a task to \code{working} (Section~\ref{sec:progress}). The field \code{reason} is OPTIONAL. Its value SHOULD start with a short machine-readable code in lowercase kebab-case, which MAY be followed by a colon and a brief neutral description, e.g.\ \code{clarification-needed: time period}. The same format applies to every \code{reason} field in ASP. A \code{reason} MUST NOT contain sensitive data (see Section~\ref{sec:security}).

\begin{lstlisting}
{
  "type": "state_change",
  "agentId": "urn:agent:example:research-assistant",
  "dimension": "task",
  "from": "working",
  "to": "input-required",
  "taskId": "t-8812",
  "contextId": "ctx-301",
  "reason": "clarification-needed: time period",
  "ts": "2026-09-22T17:02:11Z"
}
\end{lstlisting}

\subsection{floor\_change}

Sent on transitions in a multi-party conversation. The fields \code{conversationId}, \code{from} and \code{to} are REQUIRED; the field \code{role} is OPTIONAL (Section~\ref{sec:floor}). An agent MAY participate in several conversations at once; state is tracked per \code{conversationId}.

\begin{lstlisting}
{
  "type": "floor_change",
  "agentId": "urn:agent:example:research-assistant",
  "conversationId": "c-44",
  "from": "requesting-floor",
  "to": "has-floor",
  "role": "participant",
  "ts": "2026-09-22T17:02:15Z"
}
\end{lstlisting}

\subsection{deregister}

Sent on clean shutdown. On receipt, the server MUST set lifecycle to \code{offline} and stop liveness monitoring for that agent.

\begin{lstlisting}
{
  "type": "deregister",
  "agentId": "urn:agent:example:research-assistant",
  "reason": "scheduled-shutdown",
  "ts": "2026-09-22T18:00:00Z"
}
\end{lstlisting}

\section{Liveness}
\label{sec:liveness}

A failed agent cannot be relied on to announce its own failure. ASP therefore separates what the agent reports about itself from what the server and other components can establish independently. Section~\ref{sec:hbtimeout} covers agents that fall silent; Section~\ref{sec:beyondhb} covers agents that keep reporting but are hung or compromised.

\subsection{Heartbeat Timeout}\label{sec:hbtimeout}

Each agent declares its own heartbeat interval at registration. The server infers liveness from that interval; no server-side polling is required. The following rules apply.

\begin{itemize}
\item An agent MUST send a \code{heartbeat} at least once per declared heartbeat interval. Any other event sent by the agent itself also counts as a heartbeat; events sent about the agent by other components do not.
\item If no event arrives within 2.5 declared heartbeat intervals, the server MUST set lifecycle to \code{unreachable} and notify observers.
\item The 2.5 multiplier tolerates one lost heartbeat plus network jitter. A server MAY make it configurable but MUST NOT set it below 2.
\item When an \code{unreachable} agent sends any event, the server MUST restore lifecycle to the value that event reports, or to \code{online} if it reports none.
\item If an \code{unreachable} agent stays silent for a server-defined expiry period (RECOMMENDED: 24 hours), the server MAY treat it as \code{offline} and discard its live state.
\item \code{unreachable} and \code{stalled} are the only states an agent MUST NOT report about itself.
\end{itemize}

On transition to \code{unreachable}, the server SHOULD keep the last known task and floor states visible but mark them as stale, since the server cannot know whether they are still true.

\subsection{Beyond Heartbeats}
\label{sec:beyondhb}

The heartbeat timeout detects an agent that has stopped, lost connectivity or crashed, because it relies on the absence of a signal rather than on a message from the agent. It does not detect two other cases: an agent whose heartbeat continues while its work has stopped, and an agent that has been compromised and reports false state. The mechanisms below address them. All are OPTIONAL in this draft; a server that implements one MUST follow the rules given for it.

\subsubsection{Progress}
\label{sec:progress}

Liveness shows that the agent's process is running; progress shows that its work is advancing. An agent that emits heartbeats from a separate thread can keep reporting \code{healthy} while its main loop is blocked. To make this visible:

\begin{itemize}
\item An agent that declares \code{stallAfterSec} MUST include \code{progressSeq} in every heartbeat and MUST increase it whenever it completes a step of work, such as a model call, a tool call or a message sent.
\item If the agent has at least one task in \code{working} or \code{retrying}, the states in which progress is expected, and \code{progressSeq} has not increased for \code{stallAfterSec}, the server MUST set health to \code{stalled} with \code{source: server} and notify observers. The first heartbeat that reports a higher \code{progressSeq} clears the state and restores the health reported by the agent. Tasks that are waiting by design, such as \code{input-required}, \code{awaiting-approval}, \code{delegated} or \code{queued}, do not trigger \code{stalled}.
\item An agent MAY declare \code{maxDurationSec} on a \code{state\_change} that moves a task to \code{working}. If the task is still non-terminal when that time has elapsed, the server MAY set it to \code{timed-out} with \code{source: server}.
\item Expected durations differ widely between agents and tasks: generating images or 3D content can take far longer than producing text. \code{stallAfterSec} and \code{maxDurationSec} SHOULD therefore be set per agent and, where useful, per task type, rather than as a single deployment-wide value. Together with \code{progressSeq} and \code{activity}, they let an observer tell a long but advancing operation from an agent waiting on others and from one that is stuck.
\end{itemize}

\subsubsection{Active Probing}
\label{sec:probe}

Active probing is the only case in which the status server initiates a connection to an agent, and it is OPTIONAL for both. An agent that supports it declares \code{healthUrl} in \code{register}: an HTTPS endpoint that the status server can reach and that answers a \code{GET} request with any 2xx status when the agent is running. Probing therefore works only for agents that accept inbound connections from the server; agents behind NAT or firewalls without inbound ports simply do not declare \code{healthUrl}. When an agent that declared \code{healthUrl} misses its heartbeat deadline, the server MAY send one \code{GET} request to that endpoint before or after setting \code{unreachable}. The probe does not restore the agent's state: it only refines the reason given to observers, \code{no-heartbeat-probe-failed} when the agent does not answer and \code{no-heartbeat-probe-ok} when it answers but has stopped reporting, which points to a fault in the reporting path rather than in the agent. Probes MUST NOT be sent more often than once per declared heartbeat interval.

\subsubsection{Independent Observation}
\label{sec:observation}

The state of an agent SHOULD NOT depend only on what the agent says about itself. Components that interact with the agent, such as a supervisor that runs its process, an OFP floor manager or convener that grants it the floor, a guardrail that screens its output or a peer that calls it, often see its failures first. Such components MAY send events about another agent, with the \code{source} field set to the closest of the values in Table~\ref{tab:common}; a convener or floor manager, for example, reports as \code{peer} or \code{supervisor} depending on its authority in the deployment, using a credential authorised for that purpose (Section~\ref{sec:security}). The following precedence rules apply.

\begin{itemize}
\item \emph{Lifecycle.} States assigned by the server (\code{unreachable}) and by a supervisor (\code{crashed}) override the agent's self-report until the agent sends a new event, as in Section~\ref{sec:hbtimeout}.
\item \emph{Health.} A \code{peer} observation, for example repeated timeouts when calling the agent, does not replace the self-reported health. The server MUST record it and expose it alongside the self-report in the snapshot, and observers SHOULD highlight any disagreement between the two. A caller that enforces its own per-call timeout SHOULD report each expiry as a \code{peer} observation with \code{reason: call-timeout}, so that the timeout becomes visible to every observer rather than only to the caller.
\item \emph{Governance.} States set by a guardrail or an operator take precedence over the agent's self-report and can be cleared only by a component with the same or higher authority. Authority is ordered as \code{operator} above \code{guardrail} above \code{self}: an operator can clear a state set by a guardrail, but not the reverse.
\end{itemize}

\subsubsection{Compromised Agents}

A compromised agent can emit well-formed heartbeats and false states indefinitely, so no liveness mechanism alone can reveal it. ASP limits the damage in three ways: its governance cannot be changed by its own reports; its self-reports are shown next to independent observations, so that a contradiction becomes visible; and, when an agent is set to \code{quarantined}, the operator MAY revoke its ingest credential. After revocation the server rejects the agent's events with \code{401}, the agent becomes \code{unreachable} in the normal way, and its last known state remains visible as stale for investigation.

\section{Transport Binding}
\label{sec:transport}

\subsection{Ingest (Agent to Server)}
\label{sec:ingest}

Table~\ref{tab:ingest} lists the ingest endpoints and Table~\ref{tab:codes} the response codes, which are standard HTTP status codes as defined in RFC~9110~\cite{rfc9110}. The server MUST ignore events whose \code{eventId} it has already processed, returning \code{202}. Events with a \code{seq} lower than the last one processed for that agent from the same sender MUST be recorded in history but MUST NOT overwrite current state.

\begin{table}[H]
\centering
\caption{Ingest endpoints.}
\label{tab:ingest}
\small
\begin{tabularx}{\textwidth}{@{}llY@{}}
\toprule
Method & Path & Purpose \\
\midrule
POST & \code{/v1/events} & Submit one event. \\
POST & \code{/v1/events:batch} & Submit an array of events, for agents buffering during brief outages. \\
POST & \code{/v1/traces} & OPTIONAL. OTLP/HTTP ingest; the server maps spans to ASP events (Section~\ref{sec:otel}). \\
\bottomrule
\end{tabularx}
\end{table}

\begin{table}[H]
\centering
\caption{HTTP status codes returned by the ingest endpoints.}
\label{tab:codes}
\small
\begin{tabularx}{\textwidth}{@{}lY@{}}
\toprule
HTTP status code & Meaning \\
\midrule
\code{202 Accepted} & Event accepted. \\
\code{400 Bad Request} & Malformed JSON or schema violation. \\
\code{401 Unauthorized} & Missing or invalid credentials. \\
\code{409 Conflict} & Event sent before \code{register}, or invalid transition (e.g.\ out of a terminal task state). \\
\code{429 Too Many Requests} & Agent exceeds the server's rate limit; retry after the \code{Retry-After} header. \\
\bottomrule
\end{tabularx}
\end{table}

\subsection{Distribution (Server to Observers)}

Table~\ref{tab:dist} lists the distribution endpoints. An observer SHOULD first fetch the snapshot, then open the stream. Each SSE message~\cite{sse} carries the ASP event as its \code{data} field and the event \code{type} as the SSE \code{event} field. The SSE \code{id} field MUST carry a server sequence number, so that observers reconnecting with \code{Last-Event-ID} receive the updates they missed. The stream MAY accept filters as query parameters, such as \code{agentId} or \code{conversationId}.

\begin{table}[H]
\centering
\caption{Distribution endpoints.}
\label{tab:dist}
\small
\begin{tabularx}{\textwidth}{@{}llY@{}}
\toprule
Method & Path & Purpose \\
\midrule
GET & \code{/v1/agents} & Snapshot of current state for all agents. \\
GET & \code{/v1/agents/\{agentId\}} & Snapshot for one agent, including open tasks and conversations. \\
GET & \code{/v1/stream} & SSE stream of state updates. \\
\bottomrule
\end{tabularx}
\end{table}

\subsection{Alternative Transports}

For deployments with thousands of agents or high event rates, bindings over WebSocket, gRPC streaming or a message broker (MQTT, NATS) MAY be defined. They MUST carry the same event schema. They are not part of this draft.

\section{Mappings to Existing Standards}
\label{sec:mappings}

These mappings let adapters generate ASP events from agents that already speak other protocols, without modifying them.

\subsection{A2A}

Table~\ref{tab:map-a2a} maps A2A elements~\cite{a2a} to ASP. A2A has no notion of floor; A2A-only agents report no floor dimension.

\begin{table}[H]
\centering
\caption{Mapping from A2A to ASP.}
\label{tab:map-a2a}
\small
\begin{tabularx}{\textwidth}{@{}lY@{}}
\toprule
A2A element & ASP \\
\midrule
\code{Task.status.state} & \code{state\_change} with \code{dimension: task}; values map one-to-one to the A2A-origin states in Table~\ref{tab:task}. \\
\code{Task.id} & \code{taskId} \\
\code{contextId} & \code{contextId} \\
Agent Card \code{name}, \code{version}, \code{skills} & \code{register} fields \code{name}, \code{version}, \code{capabilities} \\
\code{TaskStatusUpdateEvent} & Source events for an adapter observing A2A streaming traffic. \\
\bottomrule
\end{tabularx}
\end{table}

\subsection{OFP}
\label{sec:ofp}

Table~\ref{tab:map-ofp} maps OFP elements~\cite{ofp} to ASP. OFP event names follow the OFP specification at the time of writing; the mapping SHOULD be revalidated against the OFP version in use.

The OFP events \code{uninvite} and \code{bye} both end an agent's participation and map to \code{left}. The \code{reason} field distinguishes the two: \code{uninvited} for a removal decided by the convener, \code{bye} for a voluntary departure.

The OFP manifest exchange (\code{getManifests}, \code{publishManifests}) has no counterpart among ASP states. Publishing a manifest is a one-time exchange of capability information rather than a condition that persists and changes over time, and capability negotiation is a non-goal of ASP. An adapter MAY copy the capabilities declared in a published manifest into the \code{capabilities} field of \code{register}.

\subsubsection{What an Adapter Can and Cannot Infer}
\label{sec:adapterlimits}

An adapter that observes OFP traffic can derive floor states and the role of each agent in the conversation reliably, because every floor transition is an explicit event. It can also infer that an agent has joined or left, or has not spoken for some time. It cannot tell, from conversation events alone, whether a silent agent is making a long model call, waiting for a tool or a peer, or is stuck: knowing who holds the floor does not reveal what the holder is doing. Lifecycle, health, progress and \code{activity} therefore require signals from the agent's runtime, either through native ASP reporting or through an adapter that has access to the runtime rather than only to the conversation. An adapter SHOULD NOT report states it cannot observe, and SHOULD leave such dimensions unreported rather than guess them.

\begin{table}[H]
\centering
\caption{Mapping from OFP to ASP.}
\label{tab:map-ofp}
\small
\begin{tabularx}{\textwidth}{@{}lY@{}}
\toprule
OFP element & ASP \\
\midrule
Conversation id & \code{conversationId} \\
\code{requestFloor} & \code{floor\_change} to \code{requesting-floor} \\
\code{grantFloor} & \code{floor\_change} to \code{has-floor} for the recipient \\
\code{yieldFloor} & \code{floor\_change} to \code{yielded} \\
\code{revokeFloor} & \code{floor\_change} to \code{floor-revoked} \\
\code{invite} & \code{floor\_change} to \code{invited} \\
\code{uninvite} & \code{floor\_change} to \code{left}, \code{reason: uninvited} \\
\code{declineInvite} & \code{floor\_change} from \code{invited} to \code{left}, \code{reason: declined} \\
\code{bye} & \code{floor\_change} to \code{left}, \code{reason: bye} \\
\code{publishManifests} & No state change; capabilities MAY populate \code{register} \\
\code{getManifests} & No ASP equivalent \\
Convener & \code{role: convener} \\
\bottomrule
\end{tabularx}
\end{table}

\subsection{OpenTelemetry}
\label{sec:otel}

ASP identifiers and metrics reuse the OpenTelemetry GenAI semantic conventions~\cite{otelgenai} where they exist (Table~\ref{tab:map-otel}). Those conventions are still experimental, and the mapping may change with them. Recommended \code{metrics} names, all OPTIONAL, are \code{latencyMs}, \code{queueDepth}, \code{activeTasks}, \code{maxConcurrentTasks}, \code{tokensIn}, \code{tokensOut}, \code{costUsd} and \code{errorRate}.

OpenTelemetry spans alone cannot express lifecycle, health or floor states. An agent ingested only via OTLP therefore shows task state but relies on span arrival for liveness.

\begin{table}[H]
\centering
\caption{Mapping from OpenTelemetry GenAI conventions to ASP.}
\label{tab:map-otel}
\small
\begin{tabularx}{\textwidth}{@{}lY@{}}
\toprule
OpenTelemetry & ASP \\
\midrule
\code{gen\_ai.agent.id} & \code{agentId} \\
\code{gen\_ai.agent.name} & \code{name} \\
\code{invoke\_agent} span start & task \code{working} \\
\code{invoke\_agent} span end, status OK & task \code{completed} \\
\code{invoke\_agent} span end, status ERROR & task \code{failed} \\
\code{gen\_ai.usage.input\_tokens} & \code{metrics.tokensIn} \\
\code{gen\_ai.usage.output\_tokens} & \code{metrics.tokensOut} \\
\bottomrule
\end{tabularx}
\end{table}

\section{Use Case: Monitoring a Multi-Agent Debate}
\label{sec:usecase}

This section follows a short sequence of ASP messages sent while a group of agents debates an open research question in an OFP conversation. The scenario matches the dashboard in Appendix~A: a Moderator acts as convener, a Planner, a Fact Checker and a Critic take turns on the floor, and a Research Assistant works on a separate task. All agents send their events to the same status server, which forwards the resulting state to a dashboard over SSE. The Moderator and the convener are not two different things that need two names. Moderator is simply the name that this example gives to one agent, and it describes the agent's function in the application, which is to moderate the debate. Convener is the role that the same agent holds in the OFP conversation, carried in the \code{role} field of its \code{floor\_change} events (Section~\ref{sec:floor}). A name belongs to the agent and does not change, whereas a role belongs to the agent's participation in one conversation: the same agent could be the convener of one conversation and a participant in another, and any agent, whatever its name, MAY hold the convener role.

\paragraph{Registration.} Each agent registers when it begins to operate in the deployment, in this example at start-up. The message below shows the registration of the Planner in full, including the common fields that the remaining examples omit.

\begin{lstlisting}
{
  "aspVersion": "0.1",
  "type": "register",
  "eventId": "5f0c2e1a-7b3d-4c9e-9a41-2d8e6b1f0a77",
  "seq": 1,
  "agentId": "urn:agent:example:planner",
  "name": "Planner",
  "version": "2.0.1",
  "protocols": ["ofp"],
  "capabilities": ["task-decomposition", "argumentation"],
  "heartbeatIntervalSec": 10,
  "ts": "2026-09-22T16:58:00Z"
}
\end{lstlisting}

\paragraph{Taking the floor.} When the Planner wants to speak, it asks the convener for the floor through OFP and reports the transition to the status server. When the Moderator grants the floor, the Planner reports the second transition. The two events travel as two separate POST requests, six seconds apart.

\begin{lstlisting}
{
  "type": "floor_change",
  "agentId": "urn:agent:example:planner",
  "conversationId": "c-44",
  "from": "joined",
  "to": "requesting-floor",
  "role": "participant",
  "ts": "2026-09-22T17:02:09Z"
}

{
  "type": "floor_change",
  "agentId": "urn:agent:example:planner",
  "conversationId": "c-44",
  "from": "requesting-floor",
  "to": "has-floor",
  "role": "participant",
  "ts": "2026-09-22T17:02:15Z"
}
\end{lstlisting}

\paragraph{Waiting for input.} Meanwhile the Research Assistant, working on task \code{t-8812}, needs a clarification before it can continue. It reports the task transition together with a short reason, so that an operator can see why the task stopped without opening the agent's logs.

\begin{lstlisting}
{
  "type": "state_change",
  "agentId": "urn:agent:example:research-assistant",
  "dimension": "task",
  "from": "working",
  "to": "input-required",
  "taskId": "t-8812",
  "reason": "clarification-needed: which time period to cover",
  "ts": "2026-09-22T17:02:11Z"
}
\end{lstlisting}

\paragraph{Quarantine by a guardrail.} A guardrail component, for example a Sentinel Agent~\cite{gosmar2025sentinelagentssecuretrustworthy}, that screens the output of the Summary Writer detects a suspected prompt injection. It sends a governance event about the Summary Writer, authenticated with its own credential. The server accepts the event because that credential is authorised to set governance states for other agents, and it records the guardrail as the setter (Section~\ref{sec:security}). A later self-report by the Summary Writer cannot clear the state.

\begin{lstlisting}
{
  "type": "state_change",
  "agentId": "urn:agent:example:summary-writer",
  "source": "guardrail",
  "dimension": "governance",
  "from": "normal",
  "to": "quarantined",
  "reason": "suspected-prompt-injection",
  "ts": "2026-09-22T17:01:58Z"
}
\end{lstlisting}

\paragraph{Missed heartbeats.} The Web Browser declared a heartbeat interval of 10 seconds, and its last event arrived at 17:01:39. When 25 seconds pass without further events, which is 2.5 times the declared interval, the server marks the agent as unreachable (Section~\ref{sec:liveness}). This transition is generated by the server itself, so no agent sends it: it appears only on the observer side. The dashboard receives it over SSE as follows, with the JSON payload split across several \code{data} lines, which SSE joins before delivery.

\begin{lstlisting}
event: state_change
id: 48213
data: {"type": "state_change",
data:  "agentId": "urn:agent:example:web-browser",
data:  "dimension": "lifecycle",
data:  "from": "busy", "to": "unreachable",
data:  "reason": "no-heartbeat-25s",
data:  "ts": "2026-09-22T17:02:04Z"}
\end{lstlisting}

Across the whole sequence no agent needs to know that a dashboard exists, and the dashboard never polls an agent: every update travels as one small POST into the server and one SSE message out of it. The last known task of the Web Browser stays visible on the dashboard, marked as stale, until the agent sends a new event or its live state expires.

\section{Governance with Sentinel Agents}
\label{sec:sentinel}

The Sentinel Agent architecture~\cite{gosmar2025sentinelagentssecuretrustworthy} places a network of Sentinel Agents alongside the functional agents of a multi-agent system. Sentinels oversee inter-agent communication and raise alerts on suspected prompt injection, hallucination, data exfiltration, collusion or privacy breaches; a Coordinator Agent supervises policy, manages agent participation and, on the basis of those alerts, isolates or quarantines misbehaving agents. The architecture describes the roles but not a common format for alerts and decisions. Without one, each sentinel needs its own integration with the coordinator, and the decision to quarantine an agent remains invisible to the conversation manager, the dashboard and the other agents until each is told separately. This section shows how ASP provides that format. Nothing in it is required for ASP conformance.

\subsection{Roles}

In ASP terms, a Sentinel Agent reports with \code{source: guardrail} and a Coordinator Agent, which acts on an operator's behalf, with \code{source: operator}; both use credentials authorised to report on other agents (Section~\ref{sec:security}). Under the precedence rules of Section~\ref{sec:observation}, a state set by the coordinator can be cleared only by the coordinator or a human operator, and no state set by either can be cleared by the monitored agent. Sentinels and the coordinator are themselves agents, so they register, send heartbeats and report their own state like any other.

\subsection{From Alert to Quarantine}

A typical sequence has three steps. It differs from the use case of Section~\ref{sec:usecase}, in which a single guardrail quarantines the agent directly: here detection and decision are separated. First, a sentinel that detects a suspected threat sets the monitored agent to \code{under-review}, with a short reason code and, optionally, detail under \code{ext}, which never contains the offending content (Section~\ref{sec:security}):

\begin{lstlisting}
{
  "type": "state_change",
  "agentId": "urn:agent:example:summary-writer",
  "source": "guardrail",
  "dimension": "governance",
  "from": "normal",
  "to": "under-review",
  "reason": "suspected-prompt-injection",
  "ext": { "org.example.sentinel": {
    "sentinelId": "urn:agent:example:sentinel-2",
    "confidence": 0.92 } },
  "ts": "2026-09-22T17:40:12Z"
}
\end{lstlisting}

Second, the coordinator, subscribed to the SSE stream with a filter on the governance dimension, receives the event within the same stream that serves the dashboard, correlates it with alerts from other sentinels and with the agent's health and task states, and may also query the agent directly. It then decides, and sets the agent to \code{quarantined} or \code{sandboxed} with \code{source: operator}, or back to \code{normal} if the alert is not confirmed. Third, the decision takes effect across the system: the convener revokes the agent's floor, which appears as \code{floor-revoked}; orchestrators stop routing tasks to it; the operator MAY revoke its ingest credential (Section~\ref{sec:beyondhb}); and every observer sees the change at the same moment.

\subsection{State as Input to Sentinels}

ASP state is also useful to sentinels as input. Patterns in the state stream can indicate threats that the content of a single message does not reveal: an agent that reports \code{healthy} while peers record repeated \code{call-timeout} observations, a task that oscillates between \code{working} and \code{retrying}, a burst of registrations from unknown agents, or an agent that delegates an unusual number of tasks. Sentinels MAY use such patterns to raise alerts, and MAY use state to focus their scrutiny, for example on newly registered or recently degraded agents.

\subsection{Monitoring the Monitors}

Because sentinels report their own liveness, a gap in coverage is visible: a sentinel that becomes \code{unreachable} or \code{stalled} leaves the agents it watches unmonitored. A coordinator SHOULD treat such a gap as a governance event in its own right, for example by setting the affected agents to \code{sandboxed} until coverage is restored. The append-only history of governance transitions, each recorded with its source and credential, provides the audit trail that the Sentinel Agent architecture calls for in support of regulatory compliance~\cite{euaiact}.

\section{Security Considerations}
\label{sec:security}

\begin{itemize}
\item \textbf{Authentication.} Every ingest request MUST be authenticated. Bearer tokens scoped per agent are RECOMMENDED; mutual TLS MAY be used. This draft does not specify how credentials are issued, scoped, rotated or revoked; these are deployment-specific in version~0.1 (Section~\ref{sec:open}). The server MUST reject events whose \code{agentId} does not match the credential, except governance events sent by a credential explicitly authorised to set governance states for other agents, such as a guardrail component, and observations sent with a \code{source} other than \code{self} by a credential explicitly authorised to report on other agents (Section~\ref{sec:observation}).
\item \textbf{Transport.} All endpoints MUST be served over TLS in any non-local deployment.
\item \rev{SV}{\textbf{Message-level integrity.} In topologies without a dedicated status server (Section~\ref{sec:topologies}), no single component verifies credentials at ingest. Events SHOULD then be signed, for example with JSON Web Signature, so that every observer can verify who reported each state, in particular who set a governance state.}
\item \textbf{Observer access.} Snapshot and stream endpoints MUST require authorisation. State reveals which systems exist, what they are doing and when they fail, which is valuable to an attacker.
\item \textbf{Data minimisation.} Events MUST NOT carry prompts, model outputs, user data or secrets. The \code{reason} fields follow the format of Section~\ref{sec:statechange}: a short code, optionally followed by a neutral description.
\item \textbf{Spoofing.} A compromised agent could report false state, e.g.\ hiding a \code{quarantined} status. Governance states set by guardrails or operators MUST take precedence over self-reported ones, and the server MUST record who set each governance state. More generally, the server MUST record the \code{source} and the credential of every event, and SHOULD expose self-reports and independent observations separately so that a compromised agent cannot mask what others observe (Section~\ref{sec:observation}).
\item \textbf{Denial of service.} Servers SHOULD rate-limit per agent and cap event size (RECOMMENDED: 16~KB).
\item \textbf{Audit.} Servers SHOULD keep an append-only history of events for forensic analysis, with retention set by the operator.
\end{itemize}

\section{Extensibility and Versioning}
\label{sec:ext}

Vendor-specific fields MUST be placed under an \code{ext} object, keyed by a reverse-domain namespace, e.g.\ \code{"ext": \{"org.example": \{...\}\}}. Servers MUST preserve and forward unknown \code{ext} content.

Custom state values MUST use the \code{x-} prefix, e.g.\ \code{x-warming-cache}. Observers SHOULD display unknown states as-is, grouped with the nearest standard state if a hint is provided.

The field \code{aspVersion} follows semantic versioning. Minor versions add optional fields or states only. A server MUST accept events from any minor version of the major version it supports.

\section{Open Issues}
\label{sec:open}

The following questions were raised for future versions of the specification. \rev{P}{They are not exhaustive: the authors invite the community to comment on them, to challenge the choices already made in this draft, and to raise further questions.}

\begin{enumerate}
\item Should floor state be tracked for A2A multi-turn exchanges, using \code{contextId} as a pseudo-conversation?
\item Is a formal JSON Schema file part of the normative specification, or published alongside it?
\item Is a pull-mode binding worth standardising, given that adapters already cover unmodifiable agents? The single probe of Section~\ref{sec:probe} is a first step; a full binding would let the server also collect state from agents that never push.
\item Should \code{peer} observations be able to change an agent's health when several independent peers agree, and if so with what quorum?
\item Should an event for updating capabilities after registration be defined, for example when an OFP agent publishes a new manifest?
\item Should ASP define a standard credential profile, for example OAuth~2.0 client credentials with separate scopes for reporting an agent's own state, reporting on other agents and observing, so that agents, sentinels and observers from different vendors can authenticate to any conforming server?
\item \rev{SV}{Should bindings be defined for topologies without a dedicated status server (Section~\ref{sec:topologies}), such as an in-band binding for OFP or an MQTT binding, and should event signing then be REQUIRED?}
\item Should ASP define a registry of governance reason codes, for example \code{data-exfiltration} or \code{suspected-collusion}, so that alerts from different sentinel implementations can be compared (Section~\ref{sec:sentinel})?
\item Should the state vocabulary, the event layer and the behaviour layer (Section~\ref{sec:layers}) be published as three separate documents?
\item How do observers know where agents are registered, especially in multi-server environments?
\end{enumerate}

\section{Future Directions}
\label{sec:future}

Version~0.1 is a working draft. The directions below are intentions rather than commitments, and their scope and order may change with feedback from the community. \rev{P}{Contributions from implementers, protocol designers and standards bodies are welcome in each of them.} \rev{GH}{The project repository~\cite{aspgithub} is intended as the common place for this work, from the text of the specification to the reference implementation, the JSON Schema and the conformance tests.}

\emph{Reference implementation and schema.} A possible next step is a reference implementation of the status server and of the observer dashboard, which would make the behaviour of Sections~\ref{sec:liveness} and~\ref{sec:transport} testable, including the optional mechanisms of Section~\ref{sec:beyondhb}. A normative JSON Schema for each event type could support automatic validation, and a conformance test suite could let implementers check each layer of Section~\ref{sec:layers} separately.

\emph{Validation with independent adopters.} The specification would benefit from being exercised by implementations other than its authors'. One candidate for a first validation is the Playground, an interactive environment for OFP conversations that already records floor events and message routing and visualises agent states on a canvas. A small experiment could start with two or three agents, use an adapter to produce ASP events from the existing Playground events, and compare the Playground view with the ASP dashboard in scenarios such as waiting for the floor, a deliberately delayed response, a timeout, and an interruption of status reporting. Such an experiment could help clarify which states can be mapped reliably from conversation events, which need additional runtime signals, and how the meaning of individual states holds up in practice. Trials with A2A-only agents, and with agents that speak neither protocol, would be a natural way to test the claim of protocol neutrality.

\emph{Governance and security.} The integration with Sentinel Agents outlined in Section~\ref{sec:sentinel} could be explored in a simulated environment, for example by looking at how quickly a sentinel alert becomes a quarantine visible to every observer, and at whether state patterns usefully complement content analysis in detecting attacks. Experience of this kind might also inform the shared registry of governance reason codes and the treatment of peer quorum listed among the open issues (Section~\ref{sec:open}).

\emph{Scale and standardisation.} For deployments with thousands of agents, bindings over WebSocket, gRPC streaming or a message broker (Section~\ref{sec:transport}) may be worth specifying, together with guidance on aggregating state across several status servers. A common credential profile, if the open issue on authentication is settled in that direction, could also be specified, so that components from different vendors can interoperate without bespoke configuration.

\section{Conclusion}
\label{sec:conclusion}

Multi-agent systems are growing faster than the means of observing them: as agents and the links between them multiply, operators lose sight of which agents are alive, what each is doing, who is waiting on whom and which agent should no longer be trusted. ASP addresses this gap with a lightweight, protocol-neutral specification. Agents push small JSON events over HTTP and observers receive the aggregated state as soon as it changes. State is modelled along five independent dimensions (lifecycle, health, task, floor and governance), and the state vocabulary can be adopted on its own.

ASP does not rely on agents to report their own failures. Liveness is inferred by the server from missed heartbeats, progress signals separate a long operation from a stuck agent, and guardrails, supervisors, operators and peers can report on other agents, with operator and guardrail decisions taking precedence over self-reports, so that a quarantine decided by a Sentinel or Coordinator Agent is visible to every component at once. By reusing the vocabularies of A2A, OFP and OpenTelemetry, and by carrying state but never content, ASP complements existing standards and protocols rather than replacing them. It defines the state, not the agent, the framework or the vendor. \rev{P}{ASP is offered as a proposal and a starting point: the authors invite the community to review it, test it against real deployments and help shape what a common standard for agent state should become\rev{GH}{, starting from the project repository~\cite{aspgithub}}.}

\bibliographystyle{plain}
\bibliography{asp-references}

\clearpage

\section*{Appendix A: Reference Dashboard}
\label{app:dashboard}

\begin{figure}[!htbp]
\centering
\includegraphics[width=\textwidth]{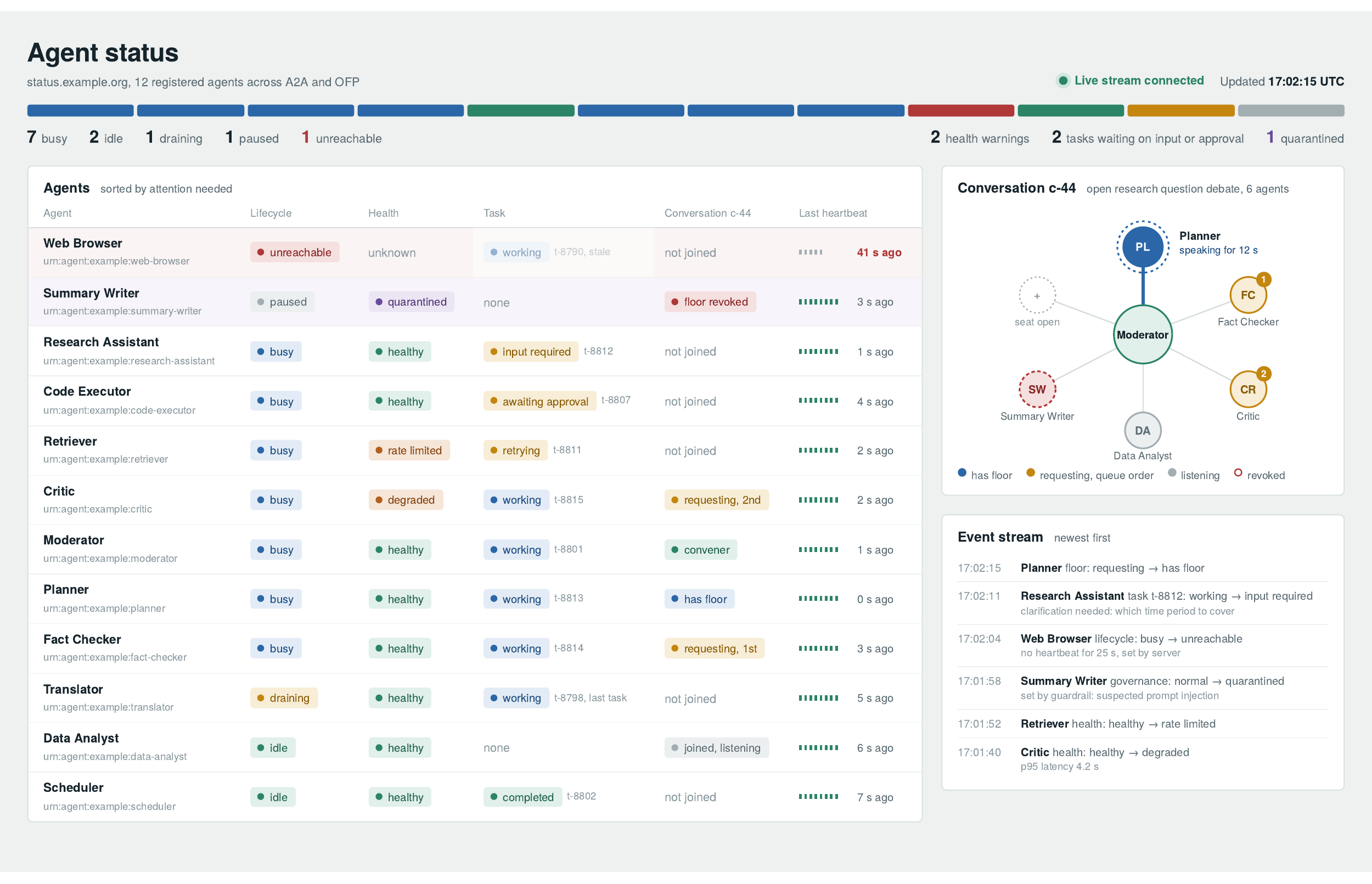}
\caption{Illustrative ASP reference dashboard: lifecycle strip and counts (top), agent table sorted by attention needed (left), floor state of an OFP conversation (upper right) and live event stream (lower right).}
\label{fig:dashboard}
\end{figure}

Figure~\ref{fig:dashboard} shows an illustrative observer dashboard built on the distribution endpoints of Section~\ref{sec:transport}. It displays twelve agents with generic roles and exercises every state dimension defined in Section~\ref{sec:states}. The dashboard is intended for a human operator who supervises the deployment; the same distribution endpoints also serve software observers.

The upper strip gives one segment per agent, coloured by lifecycle state, followed by aggregate counts. The agent table on the left is sorted by attention needed: the unreachable agent comes first, with its last known task marked as stale as required by Section~\ref{sec:liveness}, followed by the quarantined agent, whose floor has been revoked. Each row reports lifecycle, health, task and floor state, together with the age of the last heartbeat.

The panel on the upper right shows the floor state of an OFP conversation: the convener at the centre, the agent holding the floor, and the queue of agents requesting it. The event stream below lists the most recent \code{state\_change} and \code{floor\_change} events received over SSE, newest first. Every state carries a text label in addition to its colour, so the figure remains readable in greyscale. All data are illustrative.

\clearpage

\end{document}